\documentclass[10pt,twocolumn,letterpaper]{article}

\usepackage[pagenumbers]{cvpr} 

\makeatletter
\@namedef{ver@everyshi.sty}{}
\makeatother

\usepackage{graphicx}
\usepackage{float}
\usepackage{amsmath}
\usepackage{amssymb}
\usepackage{booktabs}
\usepackage{todonotes}
\usepackage{mathtools}

\usepackage{rotating}
\usepackage{array,multirow}
\usepackage{tabularx}

\usepackage[pagebackref,breaklinks,colorlinks]{hyperref}

\usepackage[capitalize]{cleveref}
\crefname{section}{Sec.}{Secs.}
\Crefname{section}{Section}{Sections}
\Crefname{table}{Table}{Tables}
\crefname{table}{Tab.}{Tabs.}

\begin{document}

\title{S$^3$Track: Self-supervised Tracking with Soft Assignment Flow}

\author{
Fatemeh Azimi$^{\star \dag \ddag}$ \quad 
Fahim Mannan$^{\dag}$ \quad
Felix Heide$^{\dag}$ \\
$^{\star}$ German Research Center for Artificial Intelligence (DFKI) \\
$^{\ddag}$ TU Kaiserslautern \quad
$^{\dag}$ Algolux \\
}

\maketitle

\begin{abstract}
In this work, we study self-supervised multiple object tracking without using any video-level association labels. 
We propose to cast the problem of multiple object tracking as learning the frame-wise associations between detections in consecutive frames. To this end, we propose differentiable soft object assignment for object association, making it possible to learn features tailored to object association with differentiable end-to-end training. With this training approach in hand, we develop an appearance-based model for learning instance-aware object features used to construct a cost matrix based on the pairwise distances between the object features.
We train our model using temporal and multi-view data, where we obtain association pseudo-labels using optical flow and disparity information.
Unlike most self-supervised tracking methods that rely on pretext tasks for learning the feature correspondences, our method is directly optimized for cross-object association in complex scenarios.
As such, the proposed method offers a reidentification-based MOT approach that is robust to training hyperparameters and does not suffer from local minima, which are a challenge in self-supervised methods.
We evaluate our proposed model on the KITTI, Waymo, nuScenes, and Argoverse datasets, consistently improving over other unsupervised methods ($7.8\%$ improvement in association accuracy on nuScenes).
\end{abstract}

\begin{figure*}[t]
\setcounter{figure}{0}
\setlength{\linewidth}{\textwidth}
\setlength{\hsize}{\textwidth}
\centering
\includegraphics[width=\linewidth]{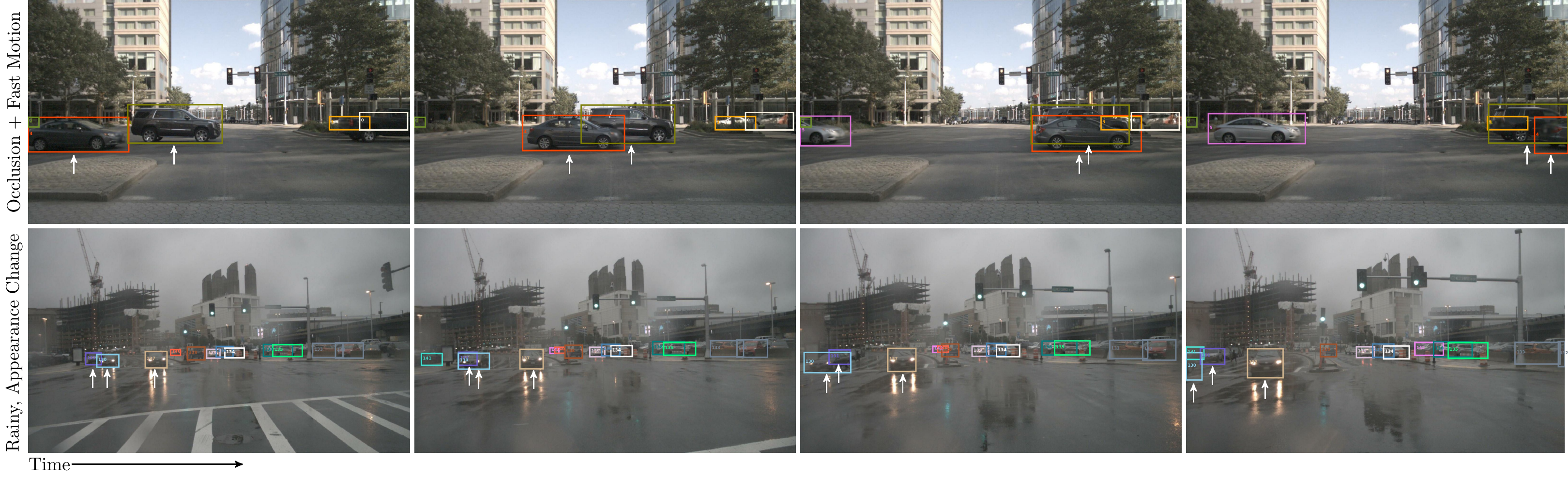}
\caption{
We propose S$^3$Track, a self-supervised method for learning the object associations throughout a video by learning a robust appearance model. We use optimal transport for computing the soft object assignments, enabling end-to-end training of our model with association pseudo-labels.
Our method shows strong performance in challenging scenarios such as occlusion and fast motion in the top row, severe weather conditions, and appearance change in the bottom row (see the objects pointed at with the white arrow).
The track IDs are visualized by the bounding box color and the number inside.
Data samples are from the nuScenes dataset \cite{caesar2020nuscenes} validation split using the provided detection bounding boxes.
}
\label{fig:intro}
\end{figure*}

\section{Introduction}
Multiple object tracking (MOT) is a fundamental task in computer vision with applications across domains, including scene understanding and autonomous driving. 
In many of these applications, tracking plays a safety-critical role for downstream planning and control algorithms. 
Accurate MOT requires precise detection of one or multiple object categories and correctly associating them throughout object presence in a dynamic scene. 
This task is challenging not only due to the similarity of object instances in the scene and highly dynamic object motion paths but also the fundamental problem of partial and full occlusions, which from the observer’s view, can break object paths into separate segments.

A large body of work has explored MOT methods during the last years \cite{wang2022recent,rakai2021data,ciaparrone2020deep,li2022learning,he2021learnable,guo2021online}, approaching the task from different viewpoints.
%
For instance, Bergman et al. \cite{tracktor_2019_ICCV} extend an object detector to a tracker using a regression network that estimates the object displacement, highlighting the importance of object detection in MOT pipelines.
In \cite{he2021learnable}, the authors improve on the standard frame-wise data association in MOT by modeling the intra-frame relationships between the tracks in the form of an undirected graph and formulating the problem as a graph-matching task.
Similarly, \cite{braso2020learning,li2020graph,wang2021joint,wang2021track,hyun2022detection}
model the interaction between objects over multiple frames as a graph and utilize graph neural networks (GNN) to globally reason over object interactions and associations.
Furthermore, recent directions that aim to tackle occlusion and association over longer time spans include architectures with attention mechanisms that equip the model with global context \cite{zeng2021motr,ma2022unified,zhou2022global,zhao2022tracking} and integrate memory to provide to utilize long-range information \cite{fu2021stmtrack,yang2018learning,cai2022memot}.

However, most MOT methods are supervised and rely on a highly laborious data labeling process, which leads to using relatively small datasets such as KITTI \cite{Geiger2012CVPR} with $21$ training sequences.
Although the recently released Waymo dataset \cite{sun2020scalability} is significantly larger than KITTI  with $798$ training videos, it is still small compared to the many hours of available unlabeled video data.
As such, this often limits the full potential of tracking architectures, as more data can significantly improve the performance of deep learning-based methods \cite{tokmakov2021learning}. 
In this work, we aim to utilize the large amount of unlabeled video data for MOT.

Object detection is closely related to MOT, as one of the main strategies for solving MOT is \textit{Tracking by Detection} which is learning to associate between the detected objects \cite{breitenstein2009robust}. 
Object detection methods are trained on separate still images;
thus, the labeling process is significantly simpler than MOT, which requires annotating sequences. 
Furthermore, the field has produced very accurate object detectors \cite{jiao2019survey,zaidi2022survey} that can be used to generate the detections for MOT. 
Although recent object detection practices utilizing transformers \cite{carion2020end,liu2021swin} show promising performance, two-stage detection methods relying on region proposals \cite{ren2015faster} are still faring among the best-performing models in a wide range of detection tasks, thanks to employing techniques such as Region of Interest (RoI) pooling and hierarchical feature processing that has been proven crucial for object detection \cite{he2015spatial,lin2017feature}.
Nevertheless, it is an open question if we can rely on the accuracy of these existing mature detection models for MOT.

In our work, we assume access to a trained object detector for generating the detection bounding boxes and train an MOT model \emph{without using video-level association labels}. 
With per-frame detections in hand, we propose a method to obtain association pseudo-labels using motion information over short video sequences.  Using the detections and the RoI pooling layer, we extract the object features and compute an affinity matrix between the detections in the source and target frames. We propose \emph{differentiable} optimal transport for finding the soft assignments between the detections, facilitating end-to-end training of our model using the association pseudo-labels. Thanks to this differentiable training, our model is able to compute robust and discriminative features optimal for object association, as shown in Figures~\ref{fig:intro} and \ref{fig:heatmap}. 
We validate the method on multiple tracking benchmark datasets, including KITTI~\cite{Geiger2012CVPR}, Argo~\cite{Argoverse2}, nuScenes~\cite{caesar2020nuscenes}, and Waymo datasets~\cite{sun2020scalability}, outperforming all tested unsupervised MOT methods.

We make the following contributions in this work:
\begin{itemize}
\setlength\itemsep{0.1em}
    \item We introduce a novel self-supervised MOT approach without video-level association annotations, hinging on a soft differentiable object assignment and object association pseudo-labels.
    \item We introduce a novel high-resolution (8MP) HDR driving dataset for multi-view self-supervised training and a pseudo-labeling method for this dataset.
    \item 
    We validate the method on the KITTI, Argo, nuScenes,
    and Waymo datasets, outperforming other unsupervised MOT methods. We confirm the effectiveness of
    all method components in ablation experiments.
\end{itemize}

\section{Related Work}
\noindent
\textbf{Multiple Object Tracking} is the task of detecting and associating multiple objects throughout a video sequence. 
MOT is an active area with different directions \cite{chaabane2021deft,chu2019famnet,zhang2019robust,braso2020learning,tracktor_2019_ICCV,hung2020soda,kim2021eagermot} that have been proposed for solving this task.
In the following, we review existing threads in MOT, followed by recent works on self-supervised techniques for spatiotemporal correspondence learning in video data.

Tracking by detection is a popular paradigm that aims to learn MOT by first detecting the objects and then finding the associations between the detections over multiple frames \cite{breitenstein2009robust,kalal2011tracking,bewley2016simple}. 
Classical methods that fall into this category mostly rely on simple motion modeling \cite{bewley2016simple,bochinski2017high} and hence, complementary to our work which utilizes visual cues.
The recent OC\_SORT method \cite{cao2022observation} makes several modifications to the Kalman-based formulation in \cite{bewley2016simple}, resulting in a considerable performance gain and better handling of occlusion. 
With the progress of deep learning methods and significantly enhanced accuracy in object detection \cite{ren2015faster,carion2020end}, this improvement has naturally carried over to tracking by detection methods \cite{bergmann2019tracking}. 
Moreover, deep networks have enabled learning of better features which improves the association accuracy \cite{wojke2017simple,Wojke2018deep}.
In CenterTrack \cite{zhou2020tracking}, a joint detection and tracking pipeline is developed for first detecting the object centers and then associating between them over consecutive frames via computing the distance between the object centers, taking the object motion offset into account.
The follow-up work PermaTrack \cite{tokmakov2021learning} utilizes the notion of physical object permanence and uses a recurrent module for memorizing the object track history and surmounting occlusion.
\cite{tokmakov2022object} further improves this method by employing a consistency-based objective for object localization in occluded videos.

Motivated by the success of transformer-based methods in vision applications \cite{vaswani2017attention,dosovitskiy2020image}, transformers have also been deployed in tracking algorithms \cite{zeng2021motr,sun2020transtrack, meinhardt2022trackformer, cai2022memot} to allow for better modeling of object relations. Sun et al. \cite{sun2020transtrack} were the first to suggest a transformer-based architecture for learning the object and track queries used for detecting objects in succeeding frames and performing association.
Trackformer \cite{meinhardt2022trackformer} proposes tracking by attention, a model which uses a transformer encoder and decoder to perform the task of set prediction between the object detections and the tracks in an autoregressive manner.
MeMOT \cite{cai2022memot} additionally utilizes an external memory for modeling the temporal context information.
In MOTR \cite{zeng2021motr}, Zeng et al. extend the deformable DETR \cite{zhu2020deformable} by building on the idea of object-to-track and joint modeling of appearance and motion by introducing a query interaction module and temporal context aggregation.

\noindent
\textbf{Self-supervised Tracking}
aims to learn spatiotemporal feature correspondences from unlabeled video data using different pretext tasks \cite{vondrick2018tracking,jabri2020space,Lai19}.
In \cite{vondrick2018tracking}, the authors propose colorization as the proxy task for learning representative features that can be used for spatiotemporal feature matching. 
Lai et al.~\cite{Lai19,lai2020mast} improve the performance of this work by employing memory in the architecture and cycle consistency during training.
Jabri et al. \cite{jabri2020space,bian2022learning} formulate the problem of correspondence learning as a contrastive random walk, while alternative approaches formulate training objectives based on
time cycle consistency \cite{wang2019learning}, or utilize motion information as the main training signal \cite{yang2021self}.
However, these algorithms developed for learning pixel-wise correspondences work well for single object tracking and often fail in crowded scenes with many similar object instances.
Moreover, the propagation-based formulation results in error accumulation for longer sequences.

Unlike dense correspondence learning, in our task, \emph{we do not require pixel-wise correspondences}, but we learn the associations between the objects.
Most similar to our work are the following two approaches: Bastani et. al \cite{bastani2021self} attempt to learn the object associates over multiple frames using a cross-input consistency objective that encourages the same association via visual and location information. 
Wang et al.~\cite{wang2020uncertainty} learn 3D object association using Kalman filtering to generate the pseudo-labels. 
Contrary to using a pretext task in \cite{bastani2021self}, we directly optimize our model on the final objective of learning to match between the detections. Unlike \cite{wang2020uncertainty}, we do not use the Kalman filter as the motion model but focus on learning the appearance model by acquiring the association pseudo-labels using motion information, specifically optical flow and disparity maps.
Finally, our work is also related to SuperGlue \cite{sarlin2020superglue}, a method for key-point correspondence search based on graph neural networks and optimal transport for feature similarity learning.
In contrast, we attempt to solve object association between consecutive frames by utilizing features extracted from object region proposals and training our model using association pseudo-labels.

\section{Tracking with Soft Assignment Flow}
\begin{figure}[t!]
    \centering
    \includegraphics[width=1.\linewidth]{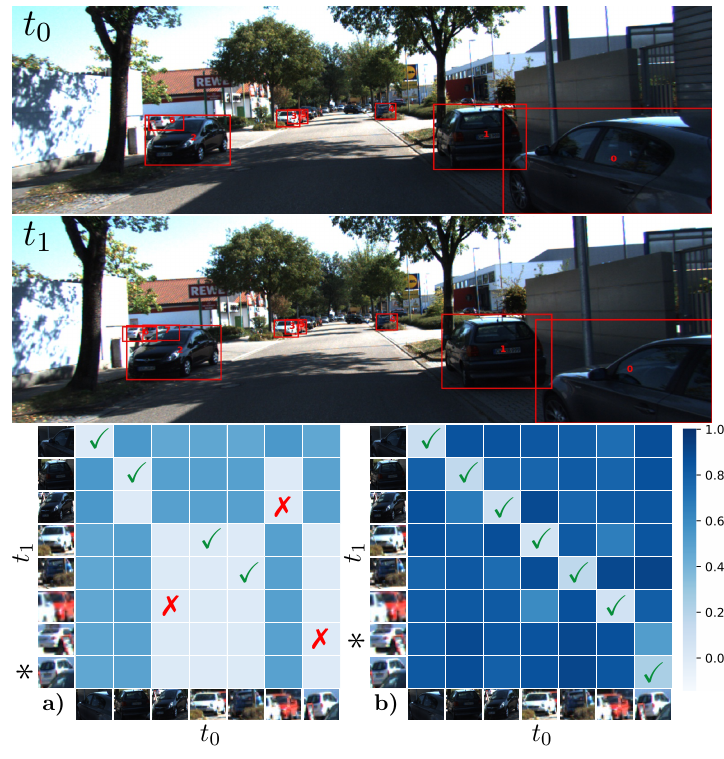}
    \caption{
        Heatmaps \textbf{a} and \textbf{b} show the cosine distance between object embeddings from an instance-agnostic model trained for object detection and our model trained for object association using optimal transport soft assignment at frames $t_0$ and $t_1$.
         The soft assignment mechanism is essential for obtaining instance-aware discriminative object features. Without this,  features from different objects are not well separated in the embedding space, resulting in a low distance between multiple object instances and false matches (\textcolor{red}{red} in heatmap \textbf{a} shows the false associations: cars 2, 5, 6, 7 at $t_1$).
         Note that our method correctly matches all detections and adequately initializes a new track ID here for car 6 entering at $t_1$. (\textbf{*}: unmatched detection resulting in a new track ID).
    }
    \label{fig:heatmap}
\end{figure}
\begin{figure*}[t!]
  \centering
  \includegraphics[width=0.9\textwidth]{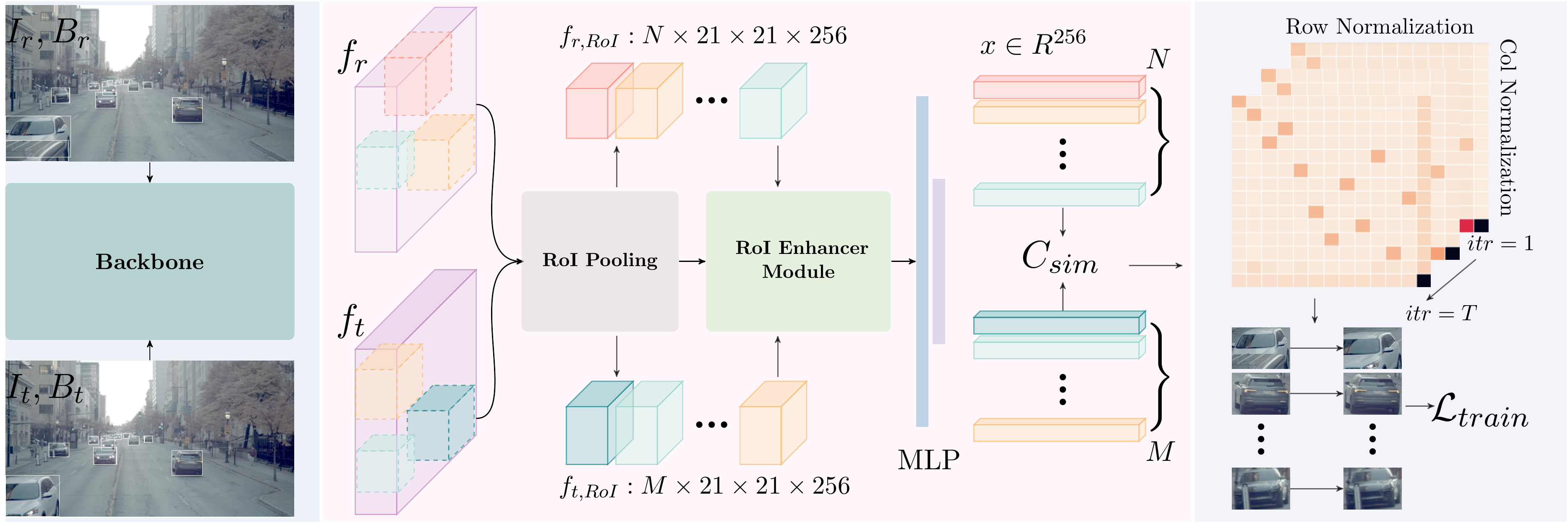}
  \caption{
        The S$^3$Track architecture.
        We use a feature pyramid pooling network as the backbone and an RoI pooling layer for extracting the per-object features $f_{RoI}$, followed by an RoI Enhancer Module generating the instance-specific discriminative representation for each object. We compute the final embeddings $x_i$ using an MLP  and find the soft assignments between the embeddings using the differentiable optimal transport layer.
        }
  \label{fig:model}
\end{figure*}
%
%
In conventional supervised MOT, a model is trained to predict a unique track ID for each object by minimizing a classification loss.
In our setting, we do not have access to the track IDs; instead, we approach MOT as finding the frame-wise association between detected objects in the reference and target frames.
Our goal is to learn an affinity matrix measuring the distance between detected objects in a reference ($I_r$) and target video image ($I_t$), using which we predict the unique correspondences between the objects such that objects with the closest distance are matched.
At first glance, this resembles the common inference strategy in MOT approaches that formulate a distance matrix based on motion or appearance information and use the Hungarian algorithm to find the unique assignments. However, the bipartite matching via the Hungarian algorithm is \emph{non-differentiable} and hence, does not facilitate learning correspondences.
To tackle this challenge, we find soft associations between the reference and target objects by posing association as an optimal transport \cite{cuturi2013sinkhorn} problem. Having defined a \emph{differentiable matching step}, we learn feature embeddings optimal for matching in an end-to-end training approach. We find this differentiable matching step essential for the proposed method, see Figure~\ref{fig:heatmap}.

We train our model with a negative log-likelihood objective where the ground truth association labels are replaced with assignment pseudo-labels obtained from video motion information.
The overall architecture is illustrated in Figure~\ref{fig:model}. In the following, we discuss all components of the proposed method.

%
%
\subsection{Optimal Transport for Soft Object Association}
We solve the task of finding the corresponding objects in $I_r$ and $I_t$ with minimal assignment distance using optimal transport \cite{cuturi2013sinkhorn,munkres1957algorithms}. We will see that this approach allows for a fully differentiable matching process.
Consider two discrete distributions of $a$ and $b$ and a matrix $C$, which represents the cost of transporting distribution $a$ to $b$ using a probability matrix $P$ (or transport matrix).
Optimal transport is a linear assignment algorithm \cite{cuturi2013sinkhorn} that finds the $P$ which minimizes the overall transport cost, that is
\begin{equation}
d_{C}(a, b) \coloneqq \min_{P \in U(a,b)} ~ \langle P,C \rangle,
\label{eq:ot}
\end{equation}
where $U(a,b)$ is the set of possible transport strategies
\begin{equation}
    U(a,b) \coloneqq \{ P \in R_{+} ~ | ~ P \mathbf{1}=a , P^{T} \mathbf{1} = b \}.
\label{eq:t_set}
\end{equation}

We are interested in finding the assignments between the detections in $I_r$ and $I_t$ such that the objects with the highest similarity (lowest distance) are matched.
In our work, we learn features optimal for object association.
Consider $X_1 \coloneqq \{ x_{1,1}, x_{1,2}, \ldots, x_{1, n_1} \}$, $X_2 \coloneqq \{x_{2,1}, x_{2,2}, \ldots, x_{2, n_2}\}$ as the set of extracted detection embeddings from the reference and target frames, where $x_{i,j} \in R^{256}$.
We define the following cost matrix for feature similarity
\begin{equation}
    C_{sim, ij} = 1 - \left\langle \frac{x_{1,i}}{\| x_{1,i} \|} , \frac{x_{2,j}}{\| x_{2,j} \|} \right\rangle
    \label{eq:feat_cost},
\end{equation}
where $\langle . , . \rangle$ represents the Frobenius dot product. 

Adding an entropy regularization term to Equation~\eqref{eq:ot} \cite{cuturi2013sinkhorn} turns the task into a convex optimization problem that can efficiently be solved using Sinkhorn algorithm \cite{sinkhorn1967concerning}.
This algorithm consists of differentiable operations, namely, iterative normalization of the rows and columns of matrix $C$ until convergence (or with a fixed number of iterations), as shown in Figure~\ref{fig:model}. With this in hand, Equation~\eqref{eq:feat_cost} allows us to \emph{learn the feature embeddings optimal for matching}.
As $n_{1}$ and $n_{2}$ may differ due to entering and exiting objects,  we augment the cost matrix $C_{sim}$ with an additional row and column, initialized with a learnable parameter $\gamma$ representing the non-match class, similar to \cite{sarlin2020superglue}. 
%
%
\subsection{Flow-based Pseudo-label Generation}
\begin{figure}[t!]
    \centering
    \includegraphics[width=\linewidth]{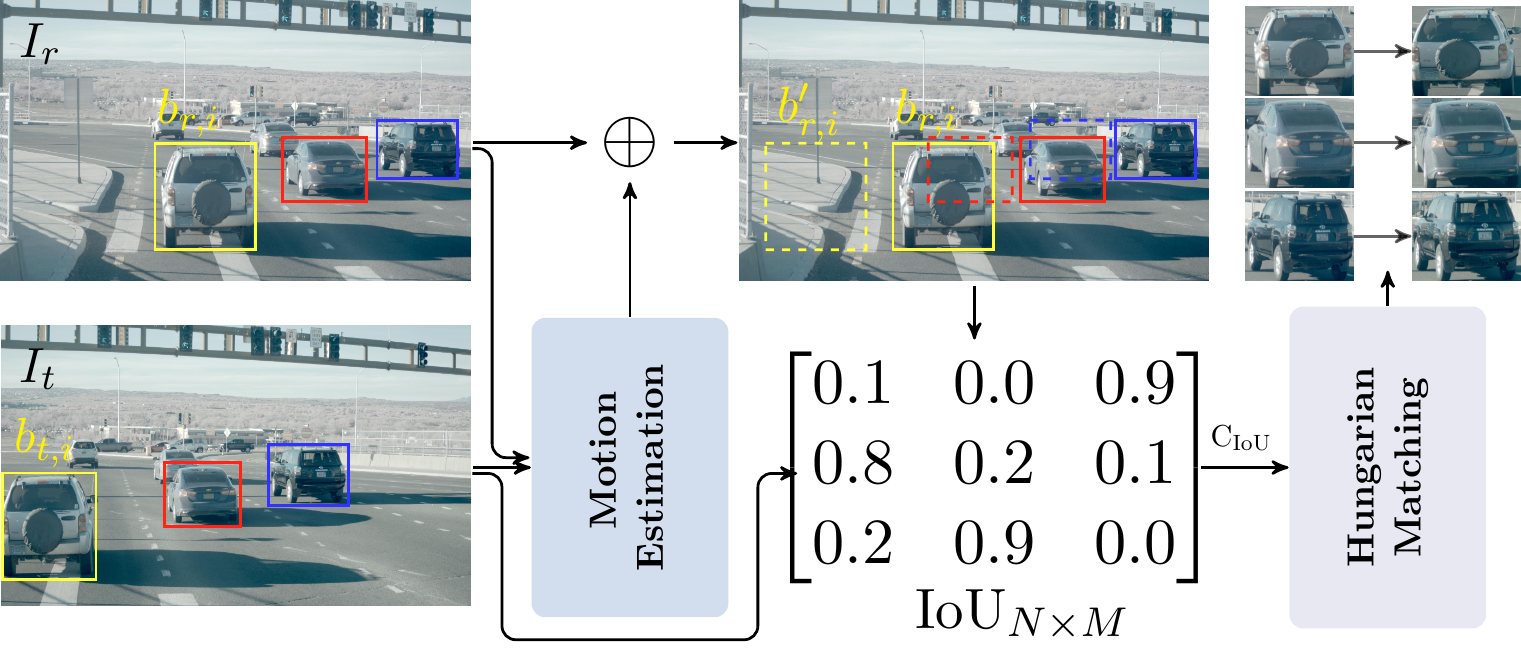}
    \caption{
    Object association pseudo-label generation process.
    We align the detection bounding boxes using motion information between a reference and a target frame.
    We compute a cost matrix based on the IoU between the aligned objects and employ the Hungarian algorithm to find the corresponding objects with maximum bounding box overlap. 
    }
    \label{fig:pseudo}
\end{figure}

\begin{figure}[tb!]
    \centering
    \includegraphics[width=\linewidth]{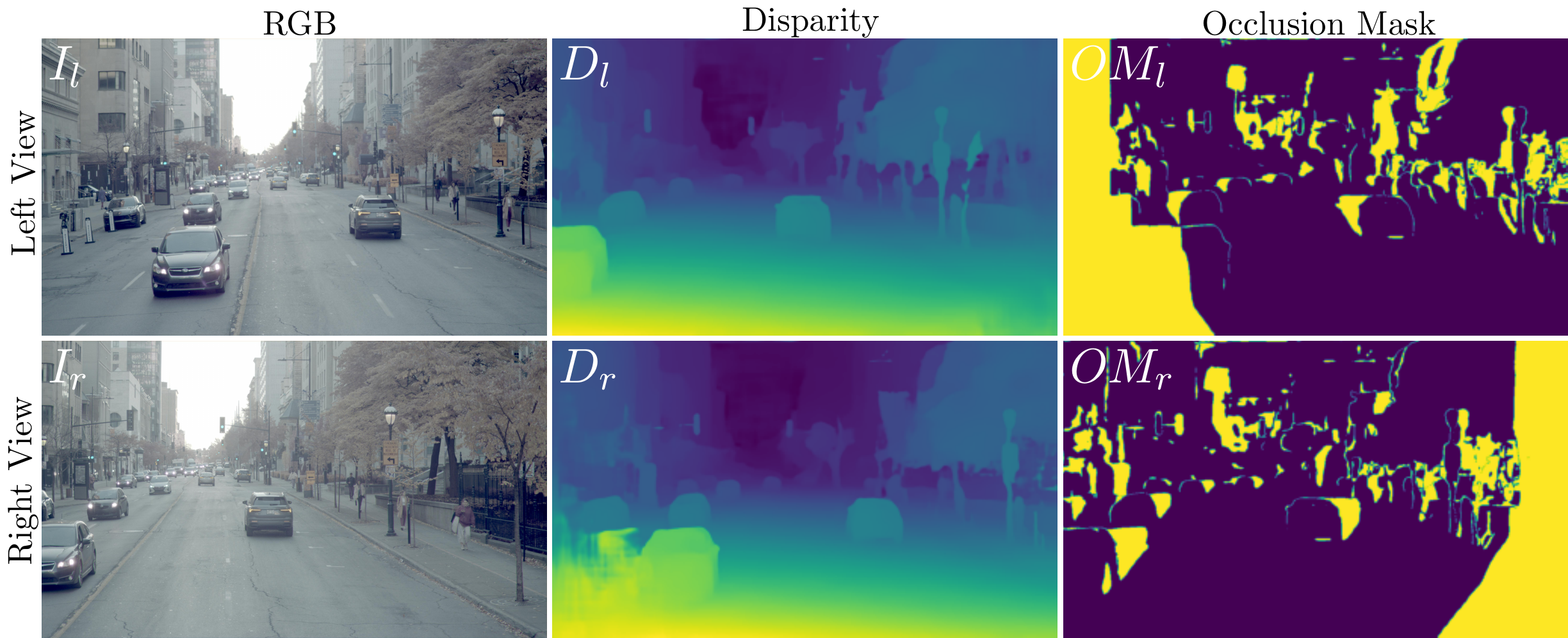}
    \caption{
    Proposed occlusion masks for the stereo data. 
    We generate $OM_l$ and $OM_r$ based on the consistency assumption that for non-occluded regions, the result of warping the left disparity $D_l$ to the right view should match $D_r$ and vice versa.
    }
    \label{fig:occ_mask}
\end{figure}

For training our association model, we recover pseudo-labels from temporal cues in video sequences and multi-view cues in stereo captures as shown in Figure~\ref{fig:pseudo}. 
To this end, we perform motion compensation to align the object bounding boxes in the reference and target frames.
We first estimate the motion between the reference and target frames $I_{r}$ and $I_{t}$.
Assume $B_r \coloneqq \{b_{r,1}, b_{r,2}, \ldots, b_{r, n_{1}} \}$ and $B_t \coloneqq \{b_{t,1}, b_{t,2}, \ldots, b_{t, n_{2}} \}$ are the detection bounding boxes in $I_{r}$ and $I_{t}$.
Then we align the detections from the reference to the target by forward warping
\begin{equation}
    b_{r,i}^{\prime} = b_{r,i} + M_{b_{i}, r}^{t} ,
\end{equation}
where $M_{b_{i}, r}^{t} $ is the computed motion vector at the center of detection box $b_{r,i}$.
In the next step, we assign pseudo-labels based on the Intersection over Union (IoU) between the motion-adjusted object bounding boxes.
Assuming aligned bounding boxes $b_t$ and $b_{r}^{\prime}$, we match objects with the highest overlap with the distance matrix
\begin{equation}
C^{\mathrm{IoU}}_{ij} = 1 - \mathrm{IoU}(b_{r,i}^{\prime}, b_{t,j}).
\end{equation}
We compute the unique and hard object association labels $(i, j)$ using the Hungarian algorithm, which gives us object correspondences with the highest overlap (minimum cost).
Note that, at this stage, we can employ the Hungarian algorithm since we do not require differentiability in the pseudo-label generation.
When using temporal data, the $I_r$ and $I_t$ are temporally spaced video frames, and motion $M$ is estimated using optical flow.
When using stereo data, $I_r$ and $I_t$  are the left and right images, and $M$ is the disparity between the two views.
\vspace{-4mm}
\paragraph{Occlusion Masks.}
Changes in camera view (from left to right stereo camera) and dynamic objects can result in occlusions. Although tracking methods should be robust to changes in appearance between frames, extreme occlusions can be detrimental to the training process.
Drastic appearance shifts and occlusion can occur for large baselines, as shown in Figure~\ref{fig:occ_mask}.
To handle this issue, we use an occlusion mask to discard objects that become heavily occluded. Specifically, we assume that for non-occluded regions, the disparity in one view should be consistent with the disparity in the other view.
We first compute the disparity maps, $
D_{l}$ and $D_r$, for the left and right views respectively.
Next, we warp $D_{l}$ to the right view, obtaining $\hat{D}_r$.
Subsequently, if the disparity difference for a pixel is above $\tau_{occ}$, that pixel is marked as occluded, that is,
\begin{equation}
    \hat{D}_{r} = \mathcal{W}(D_{l}, D_{r})
\end{equation}
\begin{equation}
    \Delta_{r} = |\hat{D}_{r} - D_{r} |, ~ OM_{r} = 
    \begin{cases}
        1 & \Delta_{r} \geq \tau_{occ} \\ 
        0 & \mathrm{otherwise}
    \end{cases},
\end{equation}
where the function $\mathcal{W}(D_{l}, D_{r})$ bi-linearly warps $D_l$ to the right view using disparity $D_r$ and $OM_{r}$ denotes the occlusion map in the right view. Similarly, we compute the occlusion mask for the left view ($OM_{l}$) and discard objects with more than $50\%$ occluded pixels in either $OM_{r}$ or $OM_{l}$ from the training data.

%
%
\subsection{Discriminative Feature Extraction}
With our differentiable assignment in hand, we train a feature extractor tailored to multi-object tracking in an end-to-end fashion. 
We find that extracting features from the detectors fails for object instance association as \emph{object detector features are instance-agnostic and not sufficiently discriminative}, as illustrated in Figure~\ref{fig:heatmap}.
To this end, we slightly modify existing object detection architectures \cite{ren2015faster} for our purpose; see Figure \ref{fig:model}.
As input, we feed the RGB image and the detection boxes from the separate detectors to the model. 
The RGB image is initially processed as a whole through a feature pyramid network \cite{lin2017feature} with ResNet50 \cite{he2016deep} backbone;
then, the RoI pooling layer extracts the context-aware object features using the detection bounding boxes. Note that this contrasts with
directly cropping the object region and then extracting the
features \cite{bastani2021self}, resulting in the complete loss of informative contextual information.
In the next step, we further process the extracted features with an RoI enhancer module consisting of a stack of convolution and non-linearity layers to obtain an instance-specific object representation specialized for the association task.
Finally, the enhanced features are projected to an embedding space $\in R^{256}$ using a small MLP network.
The resulting embeddings $x_{i}$ are used to construct the cost matrix in Equation~\eqref{eq:feat_cost}.

%
%
\subsection{Training Loss}
Assuming we have the correspondences between the detections in the reference and the target frame (from pseudo-labels), we train our model using negative log-likelihood $\mathcal{L}_{NLL}$ and triplet loss $\mathcal{L}_{trip}$, where
\begin{equation}
    \mathcal{L}_{NLL} = -\sum_{(i,j) \in \mathcal{A}} \mathrm{log}(P_{i,j}) ,
\end{equation}
with $\mathcal{A}$ being the set of association pseudo-labels between detections in $I_r$ and $I_t$.

The additional triplet loss $\mathcal{L}_{trip}$ helps in learning more discriminative features. Specifically, this loss minimizes the distance between the anchor and the positive samples and maximizes the distance between the anchor and the negative samples, up to a margin of $m$, that is
\begin{equation}
    \mathcal{L}_{trip} (a, p, n) = max\{d(a_{i}, p_{i}) - d (a_{i}, n_{i}) + m, 0\},
\end{equation}
where $a$, $p$, and $n$ stand for anchor, positive and negative samples, and $d(x_i, y_j)=|x_i - y_j|$. 
For this purpose, we select the anchor from $I_r$ and choose the positive and the negative samples from $I_t$. 
The final loss is the weighted sum of the terms above, that is
\begin{equation}
    \mathcal{L}_{train} = \alpha  \; \mathcal{L}_{trip} + \beta \; \mathcal{L}_{NLL}    
    \label{eq:lossfinal},
\end{equation}
where $\alpha$ and $\beta$ are training hyperparameters.
              
%
%
%
\subsection{Implementation Details}
We employ a feature pyramid network \cite{lin2017feature} based on ResNet50 \cite{he2016deep} as the backbone with $256$ feature channels and initialize it with pre-trained weights on the COCO dataset \cite{lin2014microsoft}.
The RoI pooling layer resizes the extracted regions to a fixed resolution of $21 \times 21$.
The RoI enhancer module consists of a 4-layer convolutional network with Group Normalization and $\mathrm{ReLU}$ non-linearity. 
The output of this block is flattened and projected to the final embedding $x \in R^{256}$ using a two-layer MLP network and $\ell_2$ normalization.
We train our model using SGD optimizer with an initial learning rate of $0.0002$, a momentum of 0.9, a weight decay of 0.0001, and a batch size of $8$ on a single A100 GPU.
We pre-train our model on temporal and multi-view driving data described in subsection~\ref{subsec:data} where we resize the images to the same width as KITTI \cite{Geiger2012CVPR}, keeping the aspect ratio unchanged. 
Additionally, we fine-tune our model on the training set of the datasets used for evaluation.
We did not find additional data augmentation beneficial to the final performance.
As training hyperparameters, we set the $\alpha$ and $\beta$ in Equation~\eqref{eq:lossfinal} to $1.0$ and $0.5$, respectively.
During inference, we define the cost matrix as the combination of appearance similarity and IoU, that is
\begin{equation}
    C_{inf} = \sigma C_{sim} + (1 - \sigma) C_{\mathrm{IoU}} .
\end{equation}
The IoU information serves as a location prior that helps the model in cases where there are similar objects present in the scene.
The relevance of IoU information highly depends on the data frame rate. 
Therefore, we use $\sigma=0.7$ and  $\sigma=1$ for the test data captured at higher (e.g., $10$ FPS) and lower (e.g., $2$ FPS) frame rates, respectively.

\vspace{2mm}
\noindent
\textbf{Pseudo-label Generation.}
We generate the initial detections per frame using a FasterRCNN meta-architecture \cite{ren2015faster} with ResNet50 as the backbone and trained on an annotated driving dataset.
To obtain more accurate pseudo-labels, we apply non-maximum suppression with an IoU threshold of $0.3$ to the detections and discard the highly overlapping bounding boxes.
We only consider detections with a minimum size of $100$ pixels and prediction confidence above $0.9$.
Furthermore, the object assignments with an IoU below $0.1$ are discarded.

\begin{figure*}[ht]
    \centering
    \includegraphics[width=\linewidth]{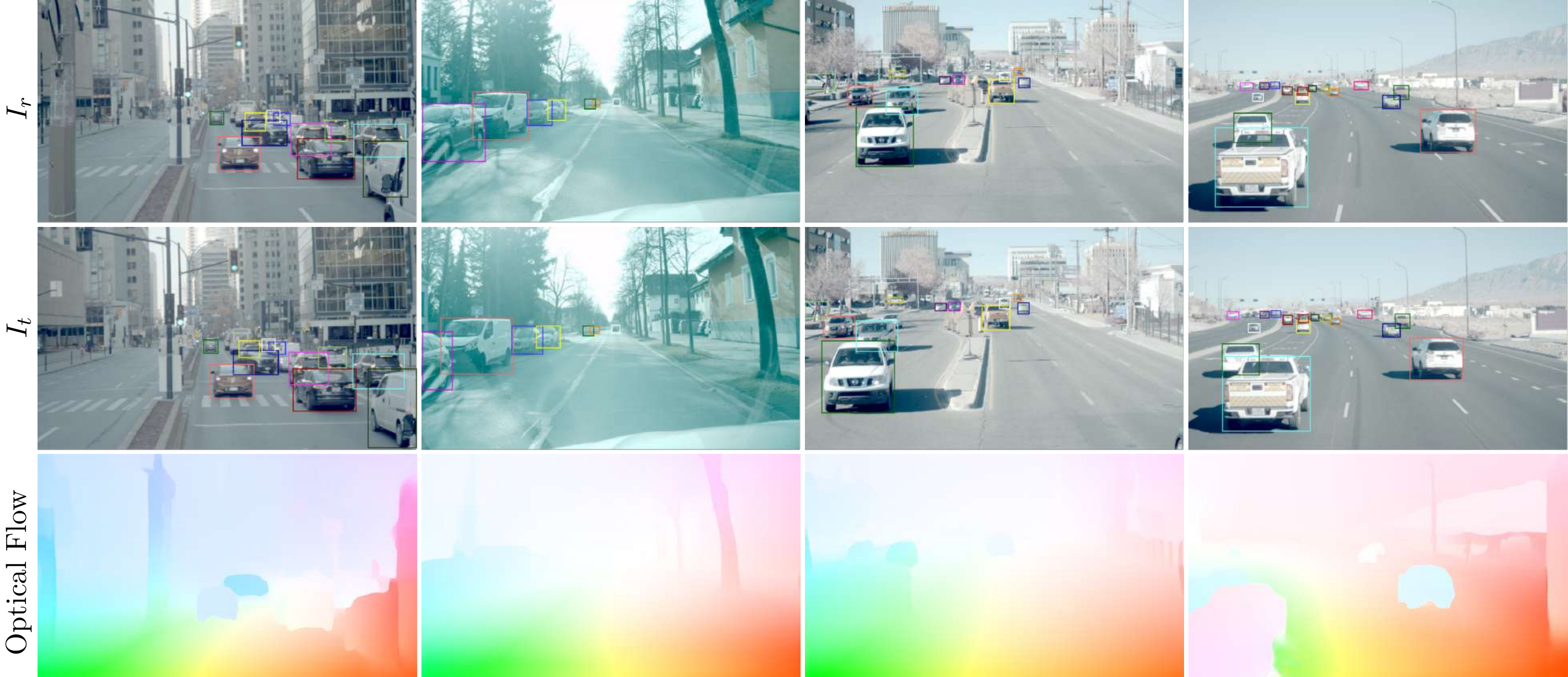}
    \caption{
    Temporal cues used during pre-training of our method.
    On our pre-training data, we use optical flow to align the detections in the reference ($I_r$) and target ($I_t$) frames.
    The association pseudo-labels are visualized by the color of bounding boxes in the first two rows.
    }
    \label{fig:train_temporal}
\end{figure*}

\begin{figure*}[t]
    \centering
    \includegraphics[width=\linewidth]{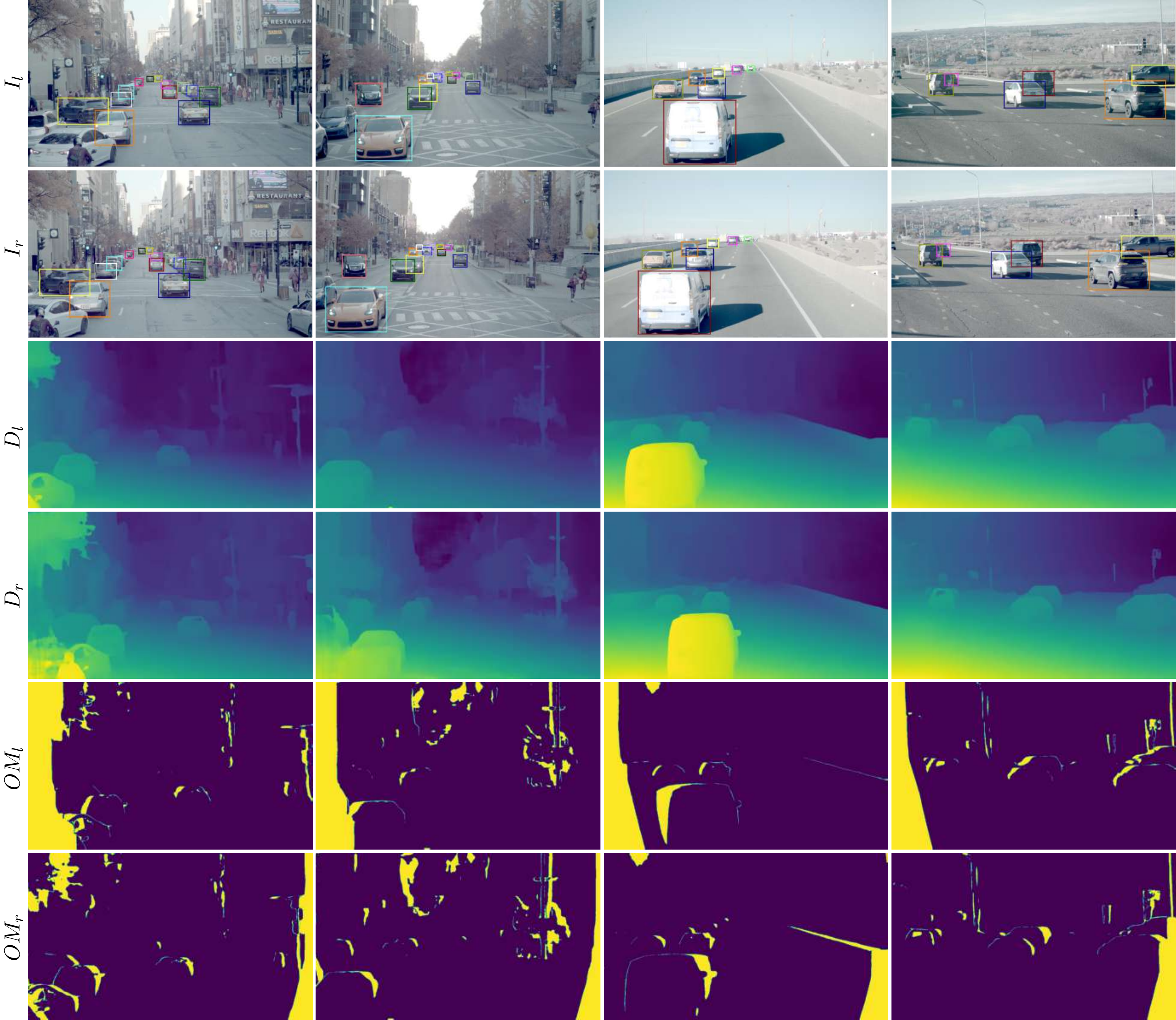}
    \caption{
    Multi-view cues ($cam0\&1$) used during pre-training.
    When using stereo data, we use disparity ($D_l$) to align the bounding boxes from the right image ($I_{r}$) to the left view ($I_l$).
    Additionally, occlusion masks $OM_l$ and $OM_r$ are utilized to discard objects that are less than 50\% visible in one of the views.
    In the first two rows, we see the left and right RGB images with the association pseudo-labels visualized with bounding box color.
    }
    \label{fig:train_stereo}
\end{figure*}

For the temporal data, we generate optical flow at $5$ FPS using the RAFT \cite{teed2020raft} model trained on the KITTI dataset \cite{Geiger2012CVPR}. 
For the stereo data, we generate the disparity using the method from Yang et al.~\cite{yang2019hierarchical}.
We generate the disparity maps from multiple stereo pairs at multiple scales and merge them to obtain a single high-resolution disparity map.
Examples of the generated pseudo-labels used in the pre-training step are shown in Figure \ref{fig:train_temporal} and Figure \ref{fig:train_stereo}. 

\section{Experiments}
\label{sec:experiments}
\noindent
In this section, we evaluate the proposed method. 
We describe the datasets, evaluation metrics, and assess S$^3$Track and relevant baselines on four autonomous driving datasets. 
We confirm our architecture choices with several ablation experiments.

%
%

\begin{table*}[t!]
\begin{center}
\small
\begin{tabular}{l|l r r r r r r r r r r}
    \toprule
    & Method & \textbf{HOTA} & DetA & \textbf{AssA} & DetRe & DetPr & \textbf{AssRe} & \textbf{AssPr} & LocA \\
    \midrule
    \multirow{4}{*}{\rotatebox{90}{\footnotesize	 \textbf{Unsup.} }}
    &SORT \cite{bewley2016simple} & 71.2 & 71.6 & 71.8 & 74.8  & 83.5  & 74.4  & 88.2  & 84.8 \\
    &IOU \cite{bochinski2017high} &74.0 & 77.4 & 71.5 & 81.2 & 85.8 &74.0 & 88.5 & 86.9 \\
    &UNS20regress\cite{bastani2021self} & 62.5 & 61.1 & 65.3 & 67.7  & 73.8  & 69.1  & 83.1  & 80.3 \\  
    &OC\_SORT \cite{cao2022observation} &76.5 & 77.3 & 76.4 & 80.6 & 86.4 & \textbf{80.3} & 87.2 & 87.0 \\
    &S$^3$Track (ours) & \textbf{76.6} & 77.5 & \textbf{76.5} & 81.3  & 85.9  & 79.6  & \textbf{88.4}  & 86.9 \\
    \midrule
    \multirow{7}{*}{\rotatebox[origin=c]{90}{\footnotesize	 \textbf{Supervised} }}
    &FAMNet \cite{chu2019famnet} & 52.6 & 61.0 & 45.5 & 64.4  & 78.7  & 48.7  & 77.4  & 81.5 \\
    &CenterTrack \cite{zhou2020tracking} & 73.0 & 75.6 & 71.2 & 80.1  & 84.6  & 73.8  & 89.0  & 86.5 \\
    &mmMOT \cite{zhang2019robust}       & 62.1 & 72.3 & 54.0 & 76.2  & 84.9 & 59.0  & 82.4  & 86.6 \\
    &LGM \cite{wang2021track}& 73.1 & 74.6 & 72.3 & 80.5  & 82.1 & 76.4  & 84.7  & 85.9 \\
    &EagerMOT \cite{kim2021eagermot} & 74.4 & 75.3 & 74.2 & 78.8 & 86.4 & 76.2 & 91.1 & 87.2 \\
    &DEFT \cite{chaabane2021deft} & 74.2 &	75.3 &	73.8 &	80.0 &	84.0 &	78.3 &	85.2 &	86.1 \\
    &PermaTrack \cite{tokmakov2021learning} & 78.0 & 78.3 & 78.4 & 81.7  & 86.5  & 81.1 & 89.5 & 87.1 \\
    &RAM \cite{tokmakov2022object} & \textbf{79.5} & 78.8 & \textbf{80.9} & 82.5 & 86.3 & \textbf{84.2} & \textbf{88.7} &	87.1 \\    
    \bottomrule
\end{tabular}
\end{center}
\caption{
        Tracking Evaluation on the KITTI test set \cite{Geiger2012CVPR}.
        In \textbf{bold}, we only show the metrics relevant for measuring the association performance. All metrics are in percentage. The proposed S$^3$Track achieves the best performance in most of the metrics in the unsupervised (Unsup.) category and fares better than most recent approaches in the supervised category.
        }
\label{tab:kitti_main}
\end{table*}

\begin{figure*}
    \centering
    \includegraphics[width=0.9\linewidth]{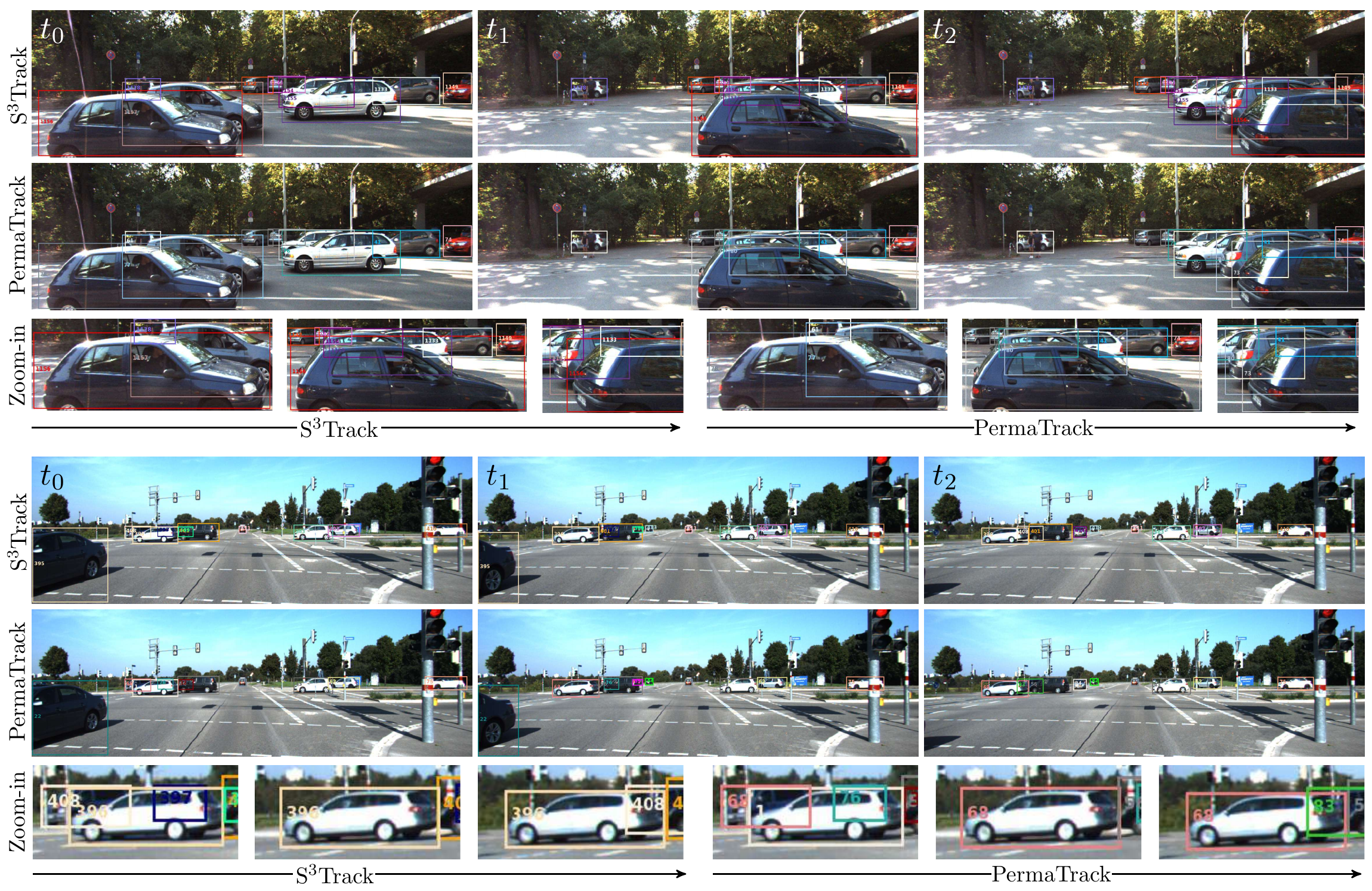}
    \caption{
    Qualitative Tracking on KITTI \cite{Geiger2012CVPR}. We compare unsupervised S$^3$Track and supervised PermaTrack \cite{tokmakov2021learning} on unseen sequences.
    The track IDs are visualized with color coding and the unique number inside each bounding box.
    Our method shows robust performance under heavy occlusion (see zoom-ins on the occluded regions).
    In both scenes, S$^3$Track correctly handles the heavy occlusion maintaining the track IDs, while PermaTrack\cite{tokmakov2021learning} suffers from several ID switches and fragmentation.
    }
    \label{fig:kitti_samples}
\end{figure*}

%
%
\begin{table*}[]
\centering
\footnotesize

\begin{tabular}{l rr  rr | rr  rr | rr  rr c}
\toprule
        \multicolumn{1}{c}{ } & \multicolumn{4}{c}{Waymo \cite{sun2020scalability}} & \multicolumn{4}{c}{nuScenes \cite{caesar2020nuscenes}}& \multicolumn{4}{c}{Argoverse \cite{Argoverse2}} &    \\
        \cmidrule(lr){2-5} \cmidrule(lr){6-9} \cmidrule(lr){10-13}

\multicolumn{1}{l}{Method}          & \multicolumn{1}{c}{AssA} & \multicolumn{1}{c}{IDF1} & \multicolumn{1}{c}{AssRe} & \multicolumn{1}{c}{AssPr} & 
                                      \multicolumn{1}{c}{AssA} & \multicolumn{1}{c}{IDF1} & \multicolumn{1}{c}{AssRe} & \multicolumn{1}{c}{AssPr} & 
                                      \multicolumn{1}{c}{AssA} & \multicolumn{1}{c}{IDF1} & \multicolumn{1}{c}{AssRe} & \multicolumn{1}{c}{AssPr} \\
                                      
                                    \midrule
SORT \cite{bewley2016simple}        & 62.2 & 71.9  & 63.9 & 93.1  & 56.5 & 66.1 & 59.2 & 82.1 & 63.2 & 75.5  & 63.9 & 93.1 \\
IOU \cite{bochinski2017high}        & 72.1 & 79.4  & 73.2 & 94.5  & 60.8 & 71.5 & 69.3 & 72.6 & 70.1 & 80.2  & 73.2 & 94.5 \\
OC\_SORT  \cite{cao2022observation} & 72.4 & 79.4  & 74.1 & 93.5  & 65.6 & 72.3 & 71.2 & 81.5 & 74.1 & 82.8  & 74.1 & 93.5 \\
S$^3$Track (ours)                   & \textbf{77.8} & \textbf{83.7} & \textbf{78.5} & \textbf{97.7}  & \textbf{73.4} & \textbf{81.9}  & \textbf{79.0} & \textbf{87.7} & \textbf{77.8} & \textbf{83.7}  & \textbf{78.5} & \textbf{93.5} \\
\bottomrule
\end{tabular}
\caption{
    Evaluation on Waymo \cite{sun2020scalability}, nuScenes \cite{caesar2020nuscenes}, and Argoverse \cite{Argoverse2} datasets.
    Our method consistently outperforms the other unsupervised methods with a considerable margin of about 4-8 points on AssA across all datasets.
    As can be seen from the results, our method shows a robust performance when processing low frame rate videos as in nuScenes dataset.
    This is contrary to the motion-based models \cite{bewley2016simple,bochinski2017high,cao2022observation}, where the association accuracy decreases at lower frame rates with an increased object motion.
}
\label{tab:waymo_nus_argo}
\end{table*}

%
%
\subsection{Datasets}
\label{subsec:data}
\begin{figure*}[ht]
    \centering
    \begin{subfigure}{.29\textwidth}
    \centering
    \includegraphics[width=0.8\linewidth]{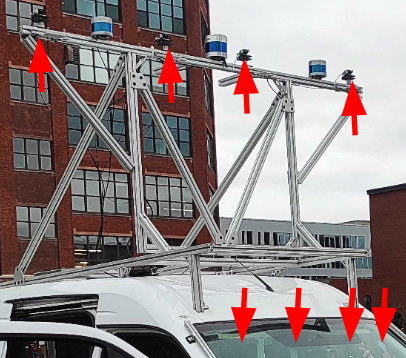}
    \caption{}
    \end{subfigure}%
    \begin{subfigure}{.75\textwidth}
    \centering
    \includegraphics[width=\linewidth]{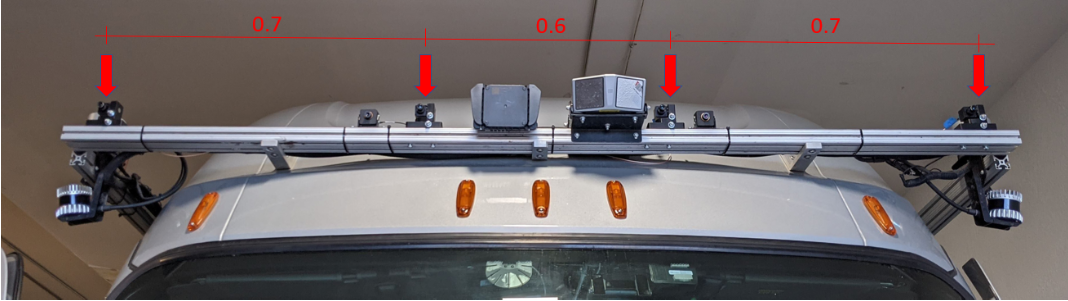}
    \caption{}
    \end{subfigure}
    \caption{4x AR0820 (8MP 20-bit HDR) cameras (top red arrows) with 30 FOV lens mounted on an $80\times20$ bar to create a large multi-view setup for data capture. We used two different vehicles (a) and (b) for data capture. For vehicle (a), data were captured at two different heights, 3m from ground and behind the windshield with a smaller baseline and about 1.5m from ground.}
    \label{fig:cam_setup}
\end{figure*}

\noindent
\textbf{Wide-baseline Stereo Pre-training Data}.
We capture a training dataset with four 8MP HDR sensors placed at 3m height on a test vehicle with baseline distance (from reference camera (${cam0}$) as 0.7m, 1.3m, and 2m for ${cam01}$, ${cam02}$, and ${cam03}$, respectively.
The primary configuration during data capture uses four front-facing 8MP 20-bit HDR sensors (AR0820) with 30 degrees horizontal field of view lenses mounted at the height of approximately 3m from the ground and distributed over a 2m baseline, as shown in \autoref{fig:cam_setup}.
During snow and rain captures, the sensors are mounted behind the windshield at a height of around 1.5m (approximate near the red arrows at the bottom of \autoref{fig:cam_setup}). 
In all cases, the cameras are mounted using a custom-made mounting plate to ensure that the cameras are attached rigidly and that there is no significant orientation difference between each pair. 
Calibration for the multi-baseline stereo was performed in two phases: lab-based offline intrinsic parameter estimation and on-site calibration using charts with clearly detectable patterns. Calibration captures were done while the vehicle was static and either in neutral or with the engine turned off to reduce any artifacts due to camera vibration and rolling shutter. 
Data capture was performed over multiple days to  collect sufficient variety in weather, illumination, and scene. A total of 52 hours of data were collected with the capture scenes, including downtown and highways, under varying illumination conditions, including noon with the sun directly above, dusk with the sun near the horizon (direct light on the sensor), and night. 
Moreover, data were collected covering clear, rainy, and snowy weather conditions.
We will release a subset of 2 hours of driving data, evenly distributed for different conditions, including data captured during \textcolor{blue}{day}, \textcolor{blue}{night}, \textcolor{blue}{dusk}, \textcolor{blue}{day+rain}, \textcolor{blue}{day+snow}, \textcolor{blue}{night+rain}, and \textcolor{blue}{night + snow}, in both \textcolor{blue}{downtown} and \textcolor{blue}{highway} traffic conditions.

\vspace{2mm}
\noindent
\textbf{KITTI} 2D tracking dataset \cite{Geiger2012CVPR} consists of $21$ training and $29$ test videos collected at $10$ FPS deploying sensors mounted on top of a driving car.
We finetune our model on the train set and evaluate it on the test set using the detections obtained from PermaTrack \cite{tokmakov2021learning}.

\noindent
\textbf{Waymo} dataset \cite{sun2020scalability} is a large-scale corpus consisting of $798$ training and $202$ validation sequences each with a duration of $20$ seconds at $10$ FPS.
We use the data captured by the front camera for fine-tuning and evaluation.

\noindent
\textbf{nuScenes} \cite{caesar2020nuscenes}
includes $700$ training videos which are annotated for 3D MOT at $2$ FPS.
Due to lower annotation frequency, this dataset has a larger appearance change compared to the KITTI and Waymo datasets.
We extract the 2D tracking labels from the 3D annotations using the scripts provided by the dataset authors and use $70$ percent of the data for finetuning and $30$ percent for validation.

\noindent
\textbf{Argoverse} dataset \cite{Argoverse2} 
also provides data for 3D tracking with a training set of $65$ and a validation set of $24$ videos.
Using the script provided by this dataset, we extracted the 2D tracking labels at $5$ FPS.
We finetune on the training data and report the performance on the validation set.

%
%
\subsection{Evaluation Metrics}
While a large set of metrics has been proposed to evaluate MOT  \cite{luiten2021hota,ristani2016performance,bernardin2008evaluating}, some existing metrics, including MOTA, are biased towards detection accuracy, hence not indicative in the context of evaluating association which is the focus of our work.
The most relevant metrics for measuring association performance are the Association Accuracy, Precision, Recall (AssA, AssPr, AssRe) \cite{luiten2021hota}, and the IDF1 score \cite{ristani2016performance}.
For completeness, we report the conventional metrics from the KITTI tracking benchmark, including Detection Accuracy, Precision, Recall (DetA, DetPr, DetRe), and the HOTA score, which combines the detection and association performance into a single number \cite{luiten2021hota}.

%
%
\subsection{Experimental Tracking Results}
We evaluate our method on the four autonomous driving benchmarks discussed above. On KITTI \cite{Geiger2012CVPR}, we compare our approach to existing supervised baselines 
and four unsupervised methods.
Like our work, the unsupervised methods do not use video-level association annotations and assume the availability of detection bounding boxes, while the \emph{supervised methods are trained using track labels}.
\cite{bewley2016simple,cao2022observation} utilize variants of Kalman filtering for modeling the object motion, while the tracker in \cite{bochinski2017high} works purely based on IoU information.
In \cite{bastani2021self}, the authors use a motion model based on bounding box information and an appearance model, and 
the self-supervised objective aims to enforce consistency between the motion and the appearance model outputs.
These methods require small object motion to work well, an assumption often violated in driving scenarios, especially with low capture frame rate.

Table~\ref{tab:kitti_main} reports tracking evaluations for the `Car' class on the KITTI \cite{Geiger2012CVPR} test set.
Together with OC\_SORT \cite{cao2022observation}, our method outperforms other unsupervised baselines and even multiple supervised methods such as CenterTrack \cite{zhou2020tracking}, EagerMOT \cite{kim2021eagermot}, and DEFT \cite{chaabane2021deft} -- which have access to track labels -- and achieves comparable results with PermaTrack \cite{tokmakov2021learning} \emph{without using any video-level association labels}.
We highlight that \cite{cao2022observation} is a purely motion-based approach and can be complementary to our proposed appearance-based method.
In Figure~\ref{fig:kitti_samples}, qualitative examples show that S$^3$Track outperforms PermaTrack in complex occlusion scenarios.

In Table~\ref{tab:waymo_nus_argo}, we discuss MOT evaluations for the `Car' category on several recent automotive datasets, namely Waymo \cite{sun2020scalability}, nuScenes \cite{caesar2020nuscenes}, and Argoverse \cite{Argoverse2}. 
The results for other baselines in Table~\ref{tab:waymo_nus_argo} are obtained using the published code from the respective authors.
The evaluations validate that our S$^3$Track performs well across all datasets and scales well to larger datasets with varying data characteristics such as weather conditions and frame rate.
This is in contrast with motion-based models \cite{cao2022observation,bewley2016simple,bochinski2017high}, where the performance considerably drops at lower frame rates.

%
%
\begin{table}[t!]
\centering
\small
\begin{tabular}{l rrrrrr}
\toprule
Method
                        & AssA & AssRe & AssPr & IDF1 \\
                        \midrule
S$^3$Track      & \textbf{96.1}   & \textbf{97.5}  & \textbf{97.7}   & \textbf{97.6}   \\
w/o RoI Pooling & 92.0    & 94.2 & 95.8    & 94.6  \\
w/o RoI Enhancer      & 92.9    & 95.1    & 96.2 & 95.3  \\
w/o soft assignment     & 83.5 & 88.2 & 90.1 & 89.1   \\
\bottomrule
\end{tabular}
\vspace{5pt}
\caption{Ablation Experiments. We confirm the effectiveness of the RoI-based feature extraction, RoI enhancing module, and soft object assignment with optimal transport.
To quantify the relevance of each component, we run an experiment without each component and report the change in the association performance.}
\label{tab:ablation_main}
\end{table}

\subsection{Ablation Experiments}
We conduct ablation experiments that validate the effectiveness of different components of our method. For all experiments, we train on the proposed wide-baseline stereo driving data and evaluate on the (now unseen) KITTI training set where the detections are available.
%
%
Table \ref{tab:ablation_main} shows the contribution of the main components of S$^3$Track.
In the first row, we assess the importance of the RoI pooling mechanism for extracting the context-aware object features by first extracting the object patches and then extracting the features using a ResNet50 network.
Next, we evaluate the effect of the RoI enhancer module.
In this experiment, we directly perform average pooling on the extracted RoI features to obtain the final embeddings ($x_i \in R^{256}$).
In the third ablation experiment, we inspect the role of the soft assignment, which enables the end-to-end training of our model.
Here, we compute the embeddings similar to the previous experiment without further training and use pre-trained object detection weights on the COCO dataset \cite{lin2014microsoft}. 

\noindent
\textbf{Impact of the Distance Function}.
For our experiments, we use cosine distance as the measure of closeness between object embeddings.
In Table \ref{tab:embed_ablation}, we provide experimental results when training the model with an alternative $\ell_2$ distance and using a matching network for predicting the similarity score (instead of using a pre-defined similarity/distance function).
The architecture of the matching network is an MLP consisting of 3 linear layers with 1024, 256, and 1 output channels, respectively (the output of the last layer is the similarity score).
We use ReLU non-linearity between linear layers.
The input to the matching network is the concatenation of different object-pair embeddings;
this network is expected to learn the function measuring the embedding similarity.
We observe that using the learnable function in the matching network under-performs the $\ell_2$ and cosine distance functions. 

\begin{table}[H]
\centering
\small
\begin{tabular}{lccccc}
\toprule
Distance Function           & AssA & AssRe & AssPr & IDF1 \\
            \hline
Cosine      & \textbf{94.8} & \textbf{96.5} & \textbf{96.8} & \textbf{96.9}   \\
$\ell_2$          & 94.4 & 96.2 & 96.6 & 96.7   \\
MLP        & 90.7  & 92.4 & 95.7 & 93.9  \\
\bottomrule
\end{tabular}
\caption{Ablation experiments evaluating the choice of the distance function.
An embedding distance using $\ell_2$ and cosine distance performs better than using a matching network (MLP) for learning the object similarity score.
The experiments are conducted using temporal data at 5 FPS.
}
\label{tab:embed_ablation}
\end{table}

%
%
\noindent
\textbf{Influence of Pre-training Cues}.
We investigate the influence of different training cues on the proposed method using the \emph{unseen} KITTI \cite{Geiger2012CVPR} training set.
Table \ref{tab:data_ablation} assesses the method when training with different data types including temporal video data, stereo data, and a combination of both.
To evaluate the influence of stereo cues, we tested with data from the three different camera pairs with varying baseline sizes, see the previous paragraph.
Training with data from $cam0\&1$ achieves better accuracy.
The larger baseline in $cam0\&2$ and $cam0\&3$ results in a higher object appearance shift between the two views, which, in this case, is detrimental to the accuracy due to the domain gap with the KITTI data used for evaluation.
Moreover, we find that the combination of temporal and stereo data is beneficial for training a better appearance model.

\begin{table}[H]
\centering
\small
\begin{tabular}{lrrrrrrrrr}
\toprule
Data           & AssA & AssRe & AssPr & IDF1 \\
            \hline
$cam0\&1$       & 95.1    & 96.7  & 97.1  & 97.1    \\
$cam0\&2$       & 94.9    & 96.1  & 97.6  & 96.7    \\
$cam0\&3$       & 93.3    & 94.9  & 96.5  & 95.5    \\
temporal        & 94.8    & 96.5  & 96.8  & 96.9   \\
temporal + $cam0\&1$  & \textbf{96.1}   & \textbf{97.5}  & \textbf{97.7}  & \textbf{97.6}  \\
\bottomrule
\end{tabular}
\caption{Ablation experiments for temporal and stereo pre-training cues. Here, the temporal pre-training data is sampled at $5$ FPS.}
\label{tab:data_ablation}
\end{table}

%
%
In Table \ref{tab:fps}, we study the impact of frame rate when using temporal pre-training cues.
We observe that a high frame rate achieves sub-optimal performance as there is not enough change in object appearance.
A very low frame rate also decreases the accuracy due to extreme appearance changes which differ from the testing data. These findings also transfer to the stereo configuration.

\begin{table}[H]
\centering
\begin{tabular}{lrrrrrrr}
\toprule
\small
 FPS           & AssA & AssRe & AssPr & IDF1 \\
            \hline
15      & 83.2    & 85.3     & 94.9 & 88.5 \\
5       & \textbf{94.8}    & \textbf{96.5} & \textbf{96.8} & \textbf{96.9}  \\
1       & 93.4    & 94.8     & 97.3 & 95.7 \\
\bottomrule
\end{tabular}
\caption{Ablation experiments that investigate the impact of frame rate for temporal pre-training data.}
\label{tab:fps}
\end{table}

In Table \ref{tab:pretrain}, we study the effect of the pre-training step on the final association performance on the KITTI \cite{Geiger2012CVPR} test set.
In S$^3$Track${^+}$, we first pre-train the model on our driving dataset and then finetune it on the KITTI training set.
In S$^3$Track${^-}$, we initialize the model with pre-trained weights on the COCO
dataset \cite{lin2014microsoft} and directly train the model on the KITTI training set.

\begin{table}[H]
\centering
\footnotesize
\begin{tabular}{lcrrrrrr}
\toprule

Method & Pre-training   &    HOTA    & AssA & AssRe & AssPr &  \\
            \hline
S$^3$Track${^+}$ & Yes &     76.6 & 76.5 & 79.6 & 88.4  \\
S$^3$Track${^-}$ & No   &    75.2 & 73.9 & 76.3 & 88.0   \\
\bottomrule
\end{tabular}
\caption{Ablation on the impact of pre-training on our driving dataset, evaluating on KITTI \cite{Geiger2012CVPR} test set.}
\label{tab:pretrain}
\end{table}

%
%
\subsection{Additional Qualitative Evaluations on Different Autonomous Driving Datasets}
\begin{figure*}[htb!]
    \centering
    \includegraphics[width=\linewidth]{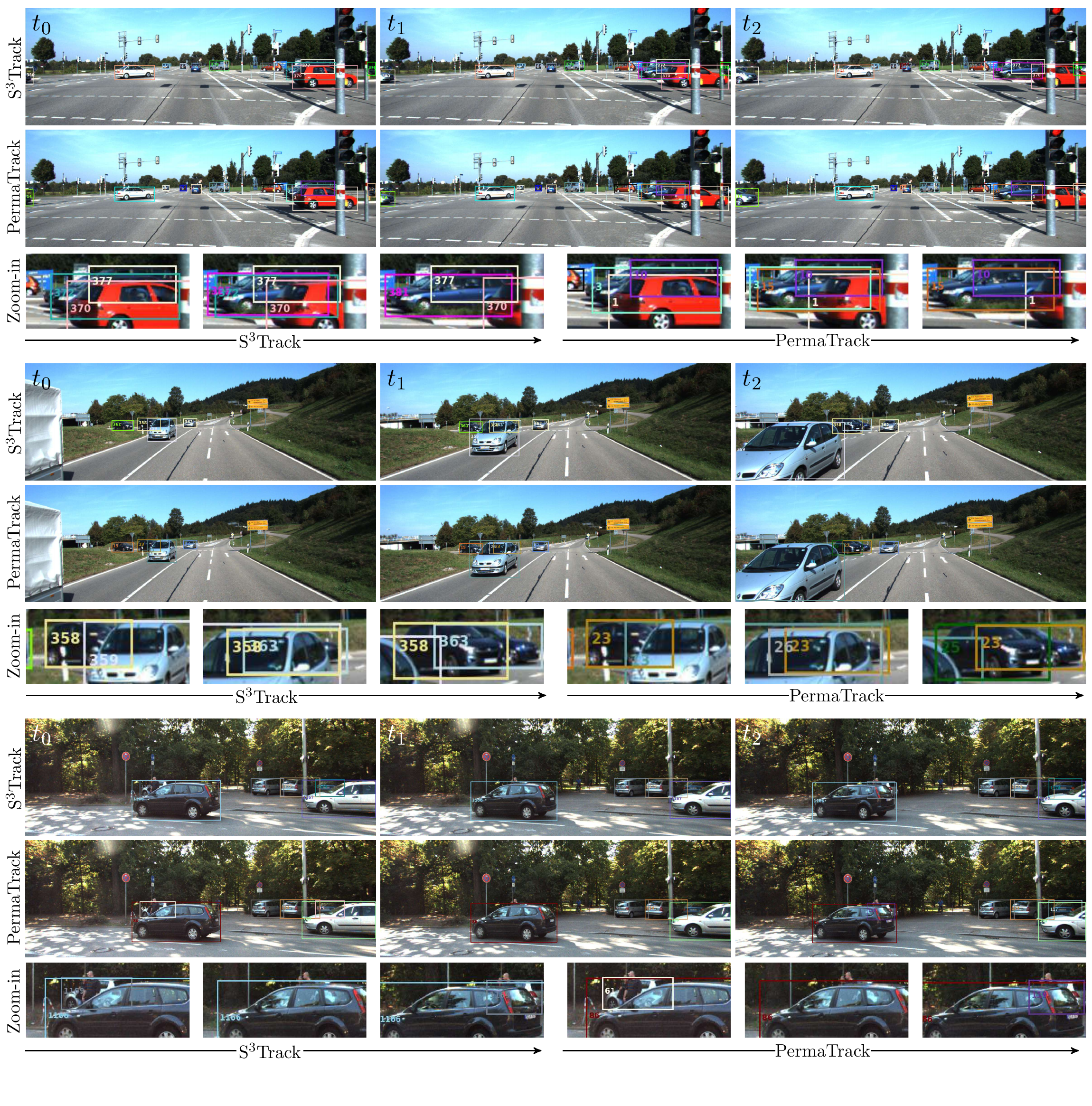}
    \caption{Additional qualitative comparisons between S$^3$Track and PermaTrack \cite{tokmakov2021learning} on KITTI \cite{Geiger2012CVPR} test set.
S$^3$Track can properly track objects throughout occlusion and is robust to missing detections.
In the first row, we see that PermaTrack suffers from an ID switch due to a missing detection, while our method preserves the correct object ID.
}
    \label{fig:kitti}
\end{figure*}

\begin{figure*}[htb!]
    \centering
    \includegraphics[width=\linewidth]{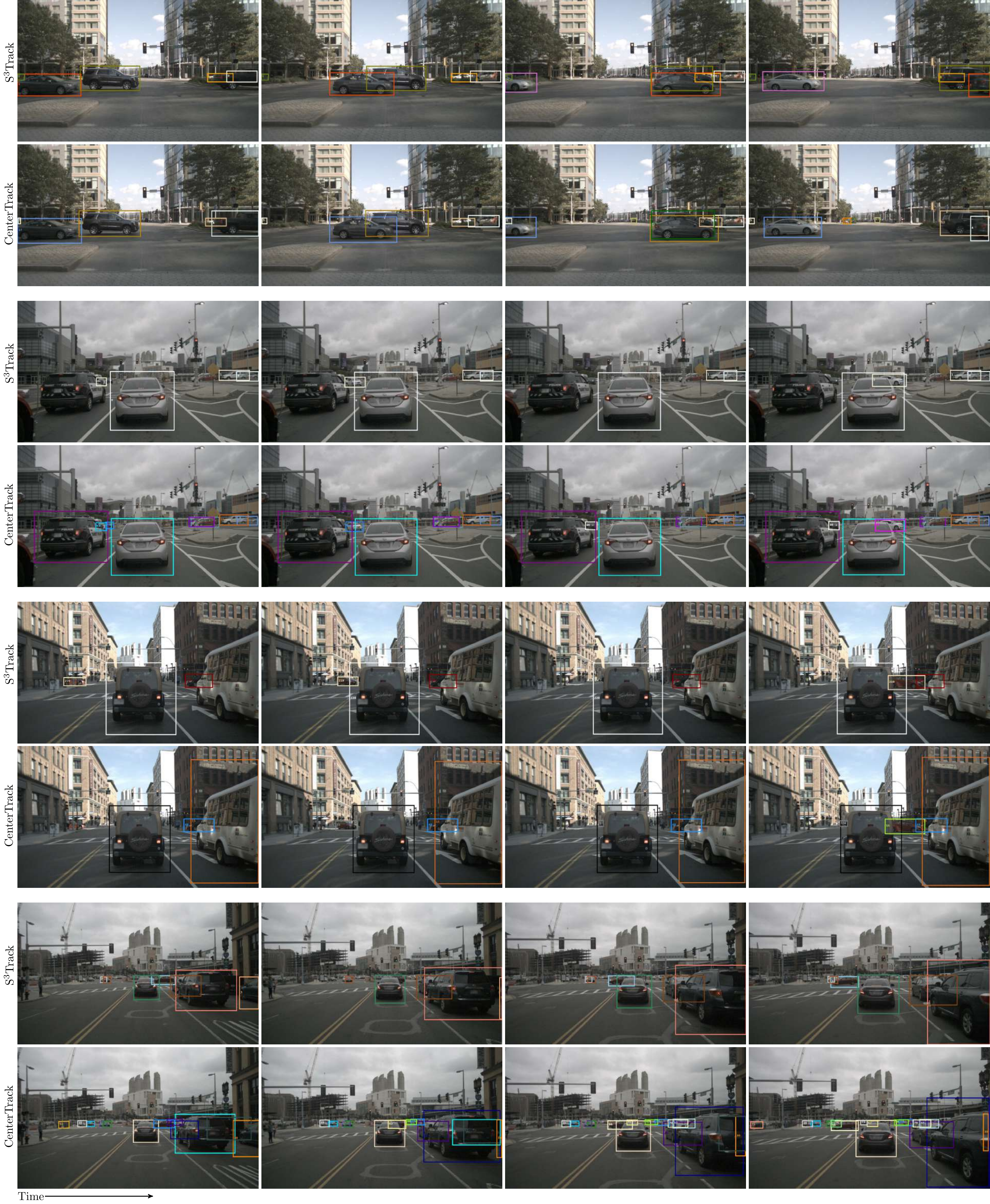}
    \caption{Additional qualitative results on nuScenes~\cite{caesar2020nuscenes}, where we compare S$^3$Track and CenterTrack \cite{zhou2020tracking}. All three examples show robustness to occlusion and large orientation change of the vehicles crossing the intersection.}
    \label{fig:nus}
\end{figure*}

\begin{figure*}[htb!]
    \centering
    \includegraphics[width=\linewidth]{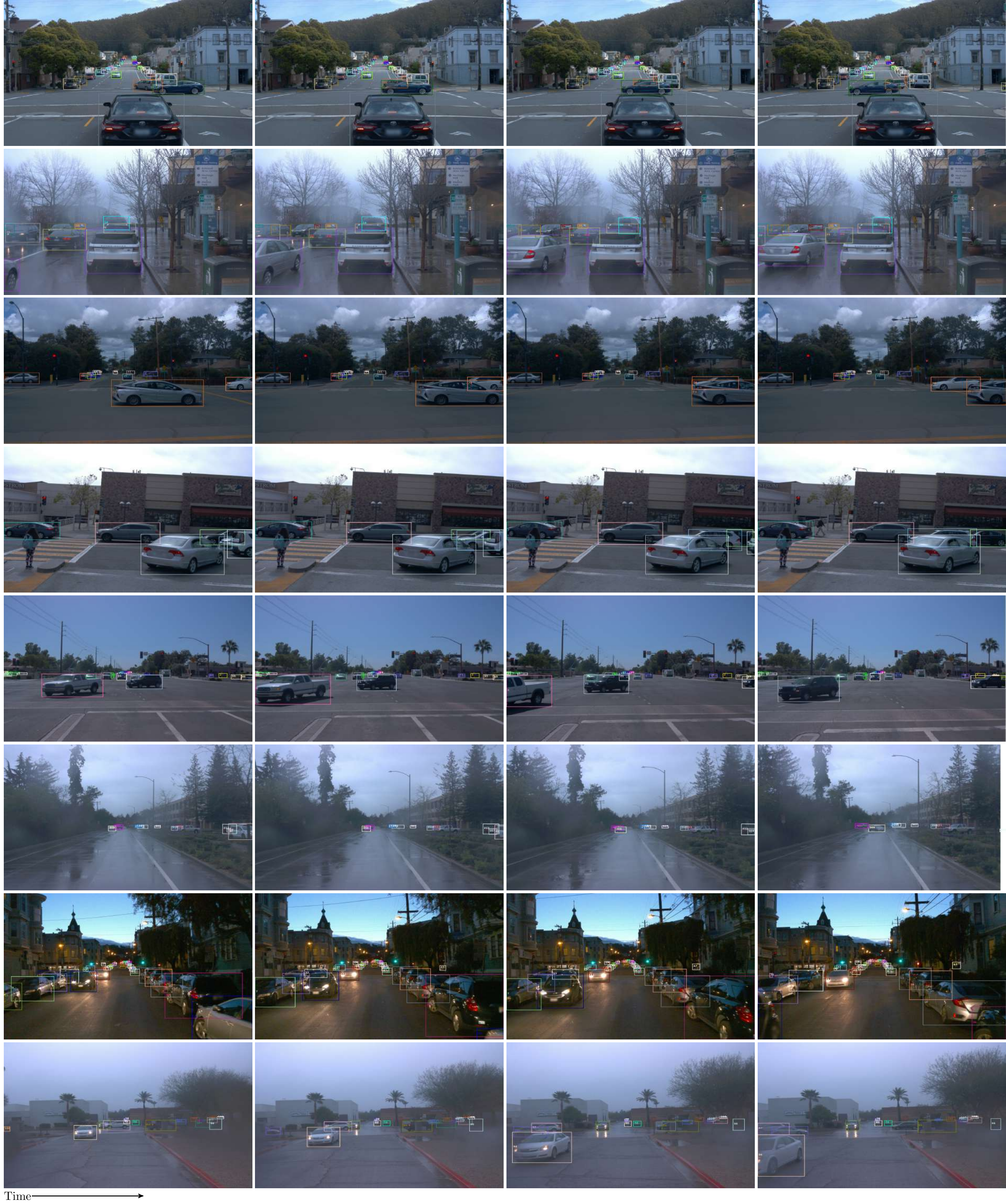}
    \caption{Additional qualitative evaluation of S$^3$Track on Waymo \cite{sun2020scalability}.
S$^3$Track is capable of tracking objects of varying sizes, can handle occlusion, and generalizes to diverse weather conditions and poor lighting.}
    \label{fig:waymo}
\end{figure*}

\begin{figure*}[htb!]
    \centering
    \includegraphics[width=\linewidth]{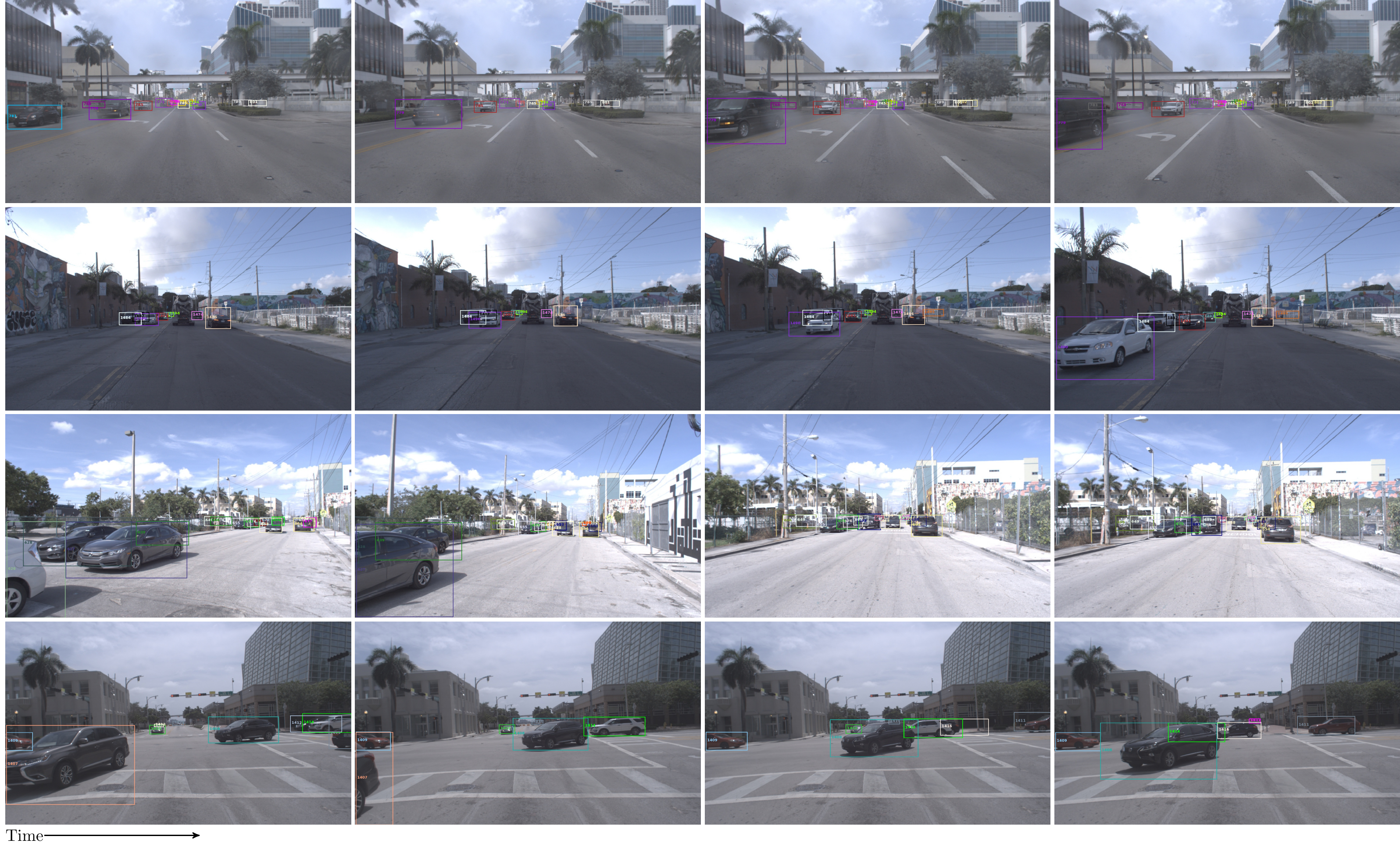}
    \caption{Additional qualitative evaluation of S$^3$Track on Argoverse \cite{Argoverse2}. The proposed method can successfully perform tracking in occluded scenes. }
    \label{fig:argo}
\end{figure*}

Next, we present and qualitative results on the KITTI \cite{Geiger2012CVPR} (Figure \ref{fig:kitti}), Waymo \cite{sun2020scalability} (Figure \ref{fig:waymo}), nuScenes \cite{caesar2020nuscenes} (Figure \ref{fig:nus}), and Argoverse \cite{Argoverse2} (Figure \ref{fig:argo}) datasets.
For the visualizations on the KITTI \cite{Geiger2012CVPR} test set, we use the detections obtained from PermaTrack \cite{tokmakov2021learning}.
For visual examples on the Waymo \cite{sun2020scalability}, nuScenes \cite{caesar2020nuscenes}, and Argoverse \cite{Argoverse2} validation splits, we use the ground-truth detections available in the respective dataset.
On nuScenes, we compare our results with CenterTrack \cite{zhou2020tracking}.
The results for CenterTrack \cite{zhou2020tracking} are generated using the code base and the trained weights published by the authors.

We illustrate challenging tracking scenarios in the qualitative results.
In Figure \ref{fig:kitti}, we see that our model is robust to missing detections (first example) and occlusion (examples 2,3) while PermaTrack undergoes multiple fragmentations and ID switches. Indeed \emph{across all datasets}, we observe robustness to occlusions with stable track IDs, see nuScenes (Figure \ref{fig:nus} samples 1 to 4), Waymo (Figure \ref{fig:waymo} rows 1,2,3,4,5), and Argoverse (Figure \ref{fig:argo} rows 1 to 4).
The experiments in Figure \ref{fig:waymo} vaidate that our model generalizes well to different weather and lighting conditions (rows 2,6,7,8), and can properly track smaller objects even when occluded (6th row).

\section{Conclusion}
We propose S$^3$Track --
a self-supervised method for multiple object tracking that operates without any video-level track labels aiming at alleviating the expensive process of data annotation for MOT.
With object bounding boxes from an accurate object detector in hand, our model performs MOT by learning the object associations over the video frames.
To this end, we propose a soft differentiable assignment approach, which allows us to train our model end-to-end using the association pseudo-labels acquired from motion information in temporal and multi-view video data. The differentiable assignment makes it possible to learn context-aware object features that are specialized for the association step. We validate our method on four autonomous driving benchmarks and demonstrate favorable performance across different datasets achieving on-par or better performance than other unsupervised methods. 
Future directions include jointly learning association and motion trajectory and exploring memory-based approaches for merging object appearance over multiple frames.

{\small
\bibliographystyle{ieee_fullname}
\bibliography{egbib}
}

\end{document}